\documentclass{article}

\usepackage{arxiv}

\usepackage[utf8]{inputenc} 
\usepackage[T1]{fontenc}    
\usepackage{mathptmx}

\usepackage{hyperref}       
\usepackage{url}            
\usepackage{booktabs}       
\usepackage{amsfonts}       
\usepackage{nicefrac}       
\usepackage{microtype}      
\usepackage{lipsum}
\usepackage{pifont}
\usepackage{graphicx}
\usepackage{pifont}
\newcommand{\cmark}{\ding{51}}%
\newcommand{\xmark}{\ding{55}}%
\usepackage{amsmath}
\usepackage{siunitx}
\graphicspath{ {./imgs/} }
\usepackage{caption}
\usepackage{subcaption}
\usepackage[super]{nth}

\usepackage{amsthm}

\title{A conflict resolution dataset derived from Argoverse-2: Analysis of the safety and efficiency impacts of autonomous vehicles at intersections}

\author{
  Guopeng Li\\
  Transport and Planning\\
  Civil Engineering and Geosciences\\
  Delft University of Technology\\
  Mekelweg 5, 2628 CD Delft \\
  \texttt{g.li-5@tudelft.nl} \\
  \And
  Yiru Jiao\\
  Transport and Planning\\
  Civil Engineering and Geosciences\\
  Delft University of Technology\\
  Mekelweg 5, 2628 CD Delft \\
  \And
  Simeon C. Calvert\\
  Transport and Planning\\
  Civil Engineering and Geosciences\\
  Delft University of Technology\\
  Mekelweg 5, 2628 CD Delft \\
  \And
  Hans (J.W.C.) van Lint \\
  Transport and Planning\\
  Civil Engineering and Geosciences\\
  Delft University of Technology\\
  Mekelweg 5, 2628 CD Delft \\
}

\begin{document}
\maketitle
\begin{abstract}
As the deployment of autonomous vehicles (AVs) in mixed traffic flow becomes increasingly prevalent, ensuring safe and smooth interactions between AVs and human agents is of critical importance. How road users resolve conflicts at intersections has significant impacts on driving safety and traffic efficiency. These impacts depend on both the behaviours of AVs and humans' reactions to the presence of AVs. Therefore, using real-world data to assess and compare the safety and efficiency measures of AV-involved and AV-free scenarios is crucial. To this end, this paper presents a high-quality conflict resolution dataset derived from the open Argoverse-2 motion forecasting data to analyse the safety and efficiency impacts of AVs. The contribution is twofold: First, we propose and apply a specific data processing pipeline to select scenarios of interest, rectify data errors, and enhance the raw data in Argoverse-2. As a result, 5000+ cases where an AV resolves conflict with a human road user and 16000+ conflict resolution cases without AVs are obtained. Motion data is smooth and consistent in these cases. This open dataset comprises diverse and balanced conflict resolution regimes. Second, this paper employs surrogate safety measures and a novel efficiency measure to assess the impact of AVs at intersections. The results suggest that human drivers exhibit similar safety and efficiency performances when interacting with AVs and with other human drivers. In contrast, pedestrians demonstrate more diverse reactions. Furthermore, due to the safety-prior strategy of AVs, the average efficiency of AV-involved conflict resolution decreases by 8.6\% compared to AV-free cases. This informative dataset provides a valuable resource for researchers and the findings give insights into the possible impacts of AVs. The dataset is openly available via \href{https://github.com/RomainLITUD/conflict_resolution_dataset}{https://github.com/RomainLITUD/conflict\_resolution\_dataset}
\end{abstract}

 
\newpage
\section{Introduction}
\label{sec: intro}

The concept of high-level Autonomous Vehicles (AVs) is expected to revolutionize urban traffic systems, offering the promise of safer roads, enhanced mobility, and increased traffic efficiency \cite{duarte2018impact}. Before reaching fully automated traffic, AVs will co-exist with human road users in the foreseeable future. In such a mixed-traffic environment, the incorporation of AVs introduces a novel dimension to the individual interactions between road agents, as well as collective traffic flow characteristics \cite{yu2021automated}. Interaction between AVs and human road users encapsulates both the control of AVs and the behavioural adjustments of human drivers. These two sides are entwined and thus necessitate a deeper understanding of how human road users re-configure their behaviours to adopt the presence of AVs. Investigating these behavioural shifts lays the groundwork for a deeper exploration of the safety and efficiency impacts that emerging AV technologies will have on ITS.

\subsection{AV-involved trajectory dataset}
\label{sec: intro data}

To comprehensively study human driver reactions and interactions with AVs, the availability of real-world AV-involved trajectory data is crucial because how humans react to the ``unhuman'' behaviours of AVs cannot be effectively modelled in a simulator. In general, the required data can be collected or generated through three methods. The first option is using driving simulators like CARLA \cite{dosovitskiy2017carla} and SVL \cite{rong2020lgsvl}. In this setup, human participants interact with designed autonomous vehicles in a virtual scenario (often facilitated by software or virtual reality equipment \cite{tran2021review}), thereby generating the data. Driving simulators have advantages in terms of safety, controllability, and standardization, making them a popular choice for safety-critical studies, such as investigations into near-collision behaviours \cite{ali2020impact} and vehicle-pedestrian interactions \cite{velasco2019studying}. The major disadvantages include limitations in physical fidelity (the so-called ``sim2real'' gap), concerns about simulation validity, and potential discomfort \cite{de2012advantages}. The second approach is field testing, where experiments are conducted in specific real-world environments, ensuring that the results accurately capture participants' naturalistic reactions. However, this method is constrained by the substantial costs associated with participant recruitment, rental of premises and equipment, and the allocation of experimental time. Additionally, ethical considerations preclude exposing vulnerable participants, such as pedestrians, to hazardous situations. Consequently, controllable field tests are typically employed to examine safe and important scenarios, such as AV-involved car-following behaviours \cite{zhao2020field}.

Recently, with the fast development of sensing technology and autonomous vehicles, employing AVs to gather trajectory data in naturalistic driving environments presents a third option. Several autonomous driving technology companies have made their road test datasets public, such as Waymo \cite{sun2020scalability}, Argoverse \cite{chang2019argoverse, wilson2023argoverse}, Lyft \cite{houston2021one}, Shifts \cite{malinin2021shifts}, and Motional Nuscenes \cite{caesar2020nuscenes}. These trajectory datasets strike a balance between realism, diversity, and scalability. However, naturalistic AV-based motion datasets lack controllability. Also, the data is very ego-vehicle-focused. This implies that one cannot directly infer the causal effect of a particular factor on safety or efficiency due to the existence of numerous confounding variables \cite{de2019causal, kumor2021sequential}. Another concern is that AV-based datasets encompass a wide range of driving scenarios, each with its own unique characteristics. It is necessary to evaluate the impacts of AVs within each distinct scenario, such as car-following and intersections, or highway and urban road conditions. Therefore, the categorization of large-scale datasets into scenario-based subsets, encompassing both AV-free and AV-involved cases, holds critical importance for the reliable assessment of safety and efficiency.

Specifically in urban traffic, current AV-involved open datasets predominantly focus on car-following behaviours on corridors. The OpenACC dataset \cite{makridis2021openacc} stands out as one of the earliest real-world field test collections. OpenACC encompasses a diverse array of vehicle models, spanning both highway and urban test campaigns, and incorporates controlled leading speed profiles to facilitate systematic studies. Hu et al. \cite{hu2022processing} process a car-following dataset from the open Waymo v1.0 motion dataset. They obtained 196 pairs of trajectories of AVs following human-driven vehicles (HVs), 274 HV-following-AV pairs, and 1032 HV-following-HV pairs, mainly on urban roads. Li et al. \cite{li2023large} extracted a larger dataset from Lyft level-5, where the AVs ran in the same city and followed a fixed route. The dataset contains over 29 thousand HV-AV pairs and over 42 thousand HV-HV pairs. These real-world car-following datasets greatly expedite research into human driving behaviour with the presence of AVs and enable a more comprehensive understanding of the potential impacts of AVs in car-following traffic.

Besides car-following, \emph{intersection} plays an equally critical role in the urban environment. Intersections serve as pivotal points for accommodating traffic flows from different directions and are often identified as efficiency bottlenecks and accident-prone areas \cite{briz2019spatial}. Compared to car-following, interactions on intersections are of higher complexity, as they are two-dimensional, multi-directional (gaming process), and comprise different types of agents (HVs, pedestrians, etc.). Compared to corridors, a significant feature of intersections is the existence of conflict areas, where traffic streams from and toward different directions intersect. Therefore, evaluating and comparing conflict resolution behaviours for AV-involved and AV-free scenarios becomes an important step towards assessing the potential safety and efficiency impacts of AVs. While several drone-based datasets like pNEUMA \cite{barmpounakis2020new}, INTERACTION \cite{zhan2019interaction}, and InD \cite{bock2020ind}, provide trajectories in interactive scenarios within purely HV environments, to the best of our knowledge, there is currently no dataset specifically focused on conflict resolution involving AVs in literature. This absence of a well-annotated, high-quality, and diverse AV-involved conflict resolution dataset hampers researchers from evaluating the potential impacts of AVs on urban traffic \cite{curtis2021knowledge}. Studying these impacts is indispensable for regulating the deployment of AVs and managing the future mixed traffic flow. Next, we will introduce the safety and efficiency assessment of AVs.

\subsection{Safety and efficiency impacts of AV}
\label{sec: intro assess}

Assessing the safety and efficiency impacts of AVs on urban traffic is essential for ensuring that this technology is deployed in a way that minimizes potential risks and disruptions to the existing traffic systems. In this subsection, we briefly overview the studies on this topic.

For safety assessment, comparing surrogate safety measures of AV-involved and AV-free cases is a common approach. Car-following is the most widely studied scenario. For example, Hu et al. \cite{hu2023autonomous} analyze the Waymo car-following dataset and observe that while AVs demonstrate superior safety performance compared to HVs, there are no significant differences for HVs when following another HV as opposed to following an AV, except for smaller jam spacing. Wen et al. \cite{wen2022characterizing}, using the same dataset, observe that human drivers exhibit reduced driving volatility, shorter time headway, and increased time-to-collision (TTC) values when following AVs than following HVs. The authors argue that the larger average TTC implies safer behaviours and lower risks of rear-end collisions when following AVs. Additionally, Wang et al. \cite{wang2023characterizing} analyze the car-following behaviours approaching stopping lines and reveal that HVs significantly adapt their behaviours to AVs when decelerating and accelerating. As for intersections, most studies leverage simulations and field experiments to study different types of interactions in depth. For instance, Reddy et al. \cite{reddy2022recognizability} employ driving simulators to investigate AV-HV interactions at T-intersections and find that the recognizability and behaviours of AVs influence human reactions. Virdi et al. \cite{virdi2019safety} conduct simulations and employ a surrogate safety assessment module to evaluate safety. Their results indicate that a low penetration rate combined with low headway can potentially reduce safety at intersections. These studies offer a range of perspectives and different findings on the safety impacts of AVs subjected to the scenario, the assumptions, and the used data.

In contrast to safety assessment, evaluating the traffic efficiency impacts of AVs encompasses multiple levels of the traffic system, ranging from the performance of a road network, and the capacity in a specific scenario, to the efficiency of one conflict resolution. Agent-based simulation and traffic flow modelling are commonly applied approaches and different efficiency measures are used. 

The efficiency of a road network with mixed traffic is usually measured by the so-called macroscopic fundamental diagram (MFD) \cite{daganzo2008analytical} and the associated maximum production (maximum weighted flow).  Lu et al. \cite{lu2020impact} is a representative example of agent-based simulation. The authors employ the Intelligent Driver Model (IDM) and use different parameter settings to describe the behaviours of AVs and HVs, respectively. Next, the MFDs are derived from the microscopic traffic simulation at different penetration rates. Under these assumptions, results show traffic efficiency improvement with the growth of AV penetration. Huang et al. \cite{huang2023characterizing} use a similar methodology but with the IDMs calibrated from real-world AV data. They also report an increase in maximum production. These studies on network-level implications of AVs typically emphasize the car-following model of AVs in the simulation. However, they often do not account for (partly) unsignalized intersections and simplify or ignore conflict resolution behaviours with different types of human agents due to the inherent complexity. This limitation could potentially impact the reliability of the simulation results.

In scenario-based assessment, the traffic efficiency is represented by the capacity on the fundamental diagram (FD), the maximum throughput, or the total travel time. For example, Calvert et al. \cite{calvert2017will} use agent-based simulation to explore the influence of vehicle automation level and the penetration rate on the performance of road corridors. The results suggest that low-level automated vehicles in mixed traffic will initially have a small negative effect on road capacities. Modelling the characteristics of AV-involved traffic flow and using conventional traffic flow theory to assess road capacity are also widely studied. This approach can also assess the capacity of intersections. Various methods, such as gap acceptance theory \cite{kimber1976capacity}, conflict techniques \cite{brilon2001capacity}, and the fictive traffic light \cite{chevallier2007macroscopic}, are proposed to derive intersection capacity. Key parameters in these methodologies, such as traffic demand and serving times for different branches of intersections, are calibrated through extensive observations of the same intersection. For an in-depth exploration of this domain, we refer readers to Yu et al. \cite{yu2021automated} for a comprehensive review. 

Assessing the microscopic-level interaction efficiency, especially the conflict resolution efficiency at intersections is important for optimizing the AV control while considering efficiency constraints. Metrics like average passing time, delay time, and loss of speed are commonly used to quantify resolution efficiency \cite{hang2022driving, zhao2023unprotected}. However, these metrics are contingent upon intersection layouts, such as the size and the number of lanes, because measuring them requires determining where the interaction or conflict resolution starts. This can be a controversial subject that triggers extensive debate. A common solution in the robotics domain is to manually fix the starting and ending points and run Monte-Carlo simulations against HVs with heterogeneous driving styles to calibrate the average passing time \cite{althoff2011comparison, wang2019trajectory}. Nevertheless, AV-based datasets are ego-vehicle-focused, which means conflicts are observed at arbitrary types of intersections. Repeated observation at the same intersection is not feasible. Additionally, these microscopic efficiency metrics cannot be related to the traffic flow properties because the influence on the continuously incoming vehicles is ignored. Therefore, we need a novel conflict resolution efficiency measure that can be applied to any intersection and facilitate the estimation of intersection capacity.

\subsection{Contributions and outline}
\label{sec: intro contribution}

In summary, the above overview highlights two critical gaps in the literature. First, there is a pressing need for an AV-involved conflict resolution dataset to facilitate the study of human responses to AVs and to assess the safety and efficiency impacts of AVs at intersections. Second, a conflict resolution efficiency metric that can be applied to intersections of varying configurations is missing in the literature. This paper aims to address these two concerns. Firstly, this paper presents a high-quality and well-labelled conflict resolution trajectory dataset derived from the open Argoverse-2 data \cite{wilson2023argoverse}. Secondly, the safety impacts of AVs in this dataset are assessed through surrogate safety measures and a new metric is proposed to assess the traffic efficiency impact of the AVs.  The major contributions of this study are summarized below:
\begin{itemize}
    \item Extract, process, enhance and classify an open AV-involved conflict resolution trajectory dataset from Argoverse-2.
    \item Analyse both the safety and efficiency impacts of autonomous vehicles on conflict resolution at intersections.
    \item Investigate whether human road users behave differently when conflicting with an AV and an HV.
\end{itemize}

The rest of this paper is organized as follows. Firstly, Section.\ref{sec: framwork} presents the methodological framework. Next, Section.\ref{Sec: Processing} describes the entire data selection, processing, and enhancement pipeline and \ref{sec: assessment} assesses the quality of the enhanced dataset and classifies the conflict regimes. By using the derived dataset, Section.\ref{sec: analysis} analyzes and compares the surrogate safety measures and conflict resolution efficiency of AV-involved and AV-free cases. Finally, Section.\ref{sec: conclusion} draws conclusions and proposes several further research directions.

\section{Methodological framework and dataset description}
\label{sec: framwork}

In this section, the methodological framework of this paper and the reasons for choosing the Argoverse-2 dataset will be first introduced.

\subsection{Framework}

\begin{figure}[h!]
\centering
\includegraphics[width=0.42\linewidth]{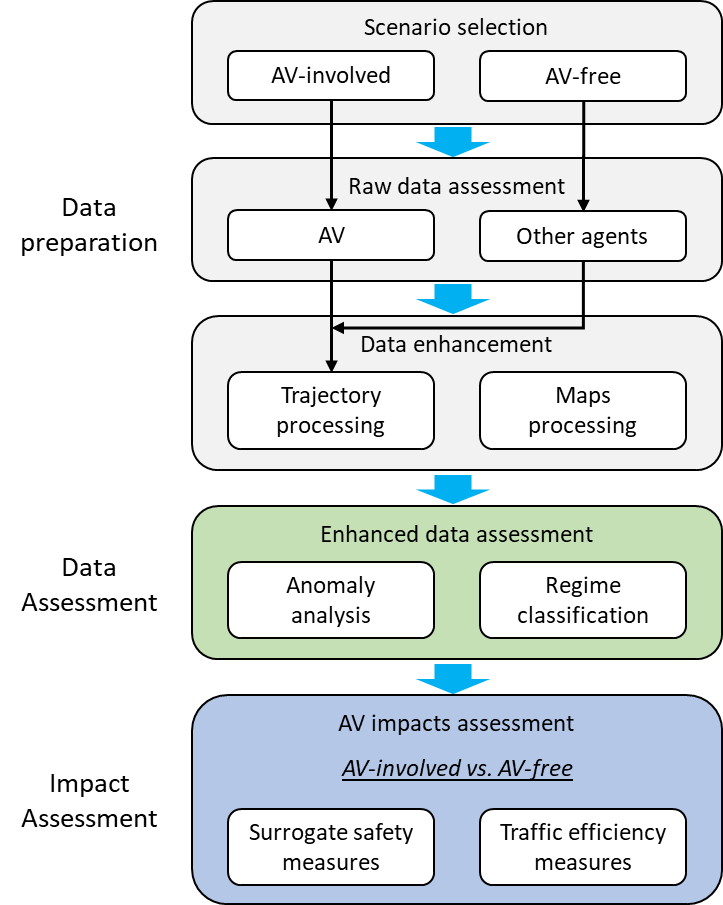}
\caption{Methodological framework of this study}
\label{fig: flowchart}
\end{figure}

Fig.\ref{fig: flowchart} shows the methodological framework of this paper. The 3 major parts are data preparation, data assessment, and impact assessment. First, a conflict resolution dataset is derived from Argoverse-2. Specifically, conflict resolution scenarios of interest (AV-involved and AV-free cases) are selected based on a set of rules. To ensure data integrity, the quality of the raw trajectories in Argoverse-2 is carefully examined to identify and rectify any existing flaws. Subsequently, a trajectory correction and smoothing pipeline is introduced and applied to enhance the raw data. Additionally, maps also undergo a re-processing step for user-friendliness. Second, a comprehensive evaluation of trajectory quality and scenario diversity within the derived dataset is undertaken. This involves anomaly analysis and regime classification. These steps serve to guarantee the robustness and representativeness of the dataset. Third, we use the enhanced dataset to study the safety and efficiency impacts of AVs. For AV-involved and AV-free cases, the surrogate safety measures and a novel traffic efficiency measure will be derived and compared to seek insights into human agents' reactions to AVs.

In the following sections, we will explain why chose the Argoverse-2 dataset and then introduce the 3 parts in more detail. 

\subsection{Argoverse-2 dataset}

To derive a conflict resolution dataset that comprises both AV-involved and AV-free scenarios, a proper data source meeting the following requirements is necessary:

\begin{itemize}
    \item AVs must be labelled. 
    \item The AV's autonomous mode must be on.
    \item The conflicts at intersections should be diverse. 
    \item The tracking duration should be long enough to cover the entire conflict resolution process.
    \item The sampling frequency should be high enough to provide behavioural details.
\end{itemize}

Table.\ref{tab: dataset compare} compares 6 popular open motion datasets. The recently-released Argoverse-2 \cite{wilson2023argoverse} motion forecasting dataset fulfils all these requirements. Argoverse-2 is collected by an automated fleet in 6 cities. The dataset contains 250k selected non-overlapping scenarios specifically focused on safety-critical long-tailed situations. The duration of each scenario is \SI{11}{\second} with a uniform 10Hz sampling rate (\SI{0.1}{\second} time interval). Elements of high-definite maps, including vectorized maps and drivable areas, are also provided. The developers also provide a Python toolkit \texttt{"av2-api"} \footnote{https://github.com/argoverse/av2-api} to read, process, and visualize the data.

\begin{table*}[h!]
    \caption{Comparison of open AV datasets}
    \label{tab: dataset compare}
    \begin{center}
    \begin{tabular}{l|cccccc}
    \toprule
    \textbf{Properties} & \textbf{Waymo}\cite{sun2020scalability}  & \textbf{Argoverse}\cite{chang2019argoverse} & \textbf{Lyft}\cite{houston2021one} & \textbf{Shifts}\cite{malinin2021shifts} & \textbf{Nuscenes}\cite{caesar2020nuscenes} & \textbf{Argoverse-2}\cite{wilson2023argoverse} \\ 
    \midrule
    AV labelled & \xmark & \cmark & \cmark & \xmark & \cmark & \cmark \\
    Autonomous mode & \cmark & ? & \cmark & ? & \xmark & \cmark \\
    Conflict diversity & \cmark & \cmark & \xmark & \xmark & \xmark & \cmark \\
    Track duration (s) & 9 & 5 & 25+ & 10 & 8 & 11\\
    Frequency (Hz) & 10 & 10 & 10 & 5 & 2 & 10\\
    \bottomrule
    \end{tabular}
    \end{center}
\end{table*}

In the first part of this paper, we aim to extract, assess, enhance, and label the AV-involved and AV-free conflict resolution cases in the Argoverse-2 motion forecasting dataset.

\section{Data preparation}
\label{Sec: Processing}

This section introduces the proposed data processing pipeline. How the dataset is extracted, corrected, and enhanced will be explained in detail.

\subsection{Scenarios selection}

Firstly, we define the use of the term ``conflict'' in this study. In general, a conflict refers to a situation in which two or more road agents create a risk of collision, disruption, or interference with each other's intended paths or movements. In this paper, we focus on one specific type of conflict: 

\emph{One vehicle (AV or HV) and another road agent with different coming lanes and different intended lanes pass the same location sequentially.} 

Most of these types of conflicts occur at intersections (including T-junctions). Lane-changing and other types of conflicts are not considered in this paper. After defining conflict, we can start by selecting conflict resolution scenarios of interest. 4 heuristic rules are applied:
\begin{itemize}
    \item The two trajectories must cross each other to exclude car-following, merging, lane-changing, etc.
    \item The two agents must have a short minimum distance or pass the same location (conflict point) in a short time interval.
    \item At least one of the two agents must change its behaviour before passing the conflict point.
\end{itemize}

\begin{figure}[h!]
\centering
\includegraphics[width=0.5\linewidth]{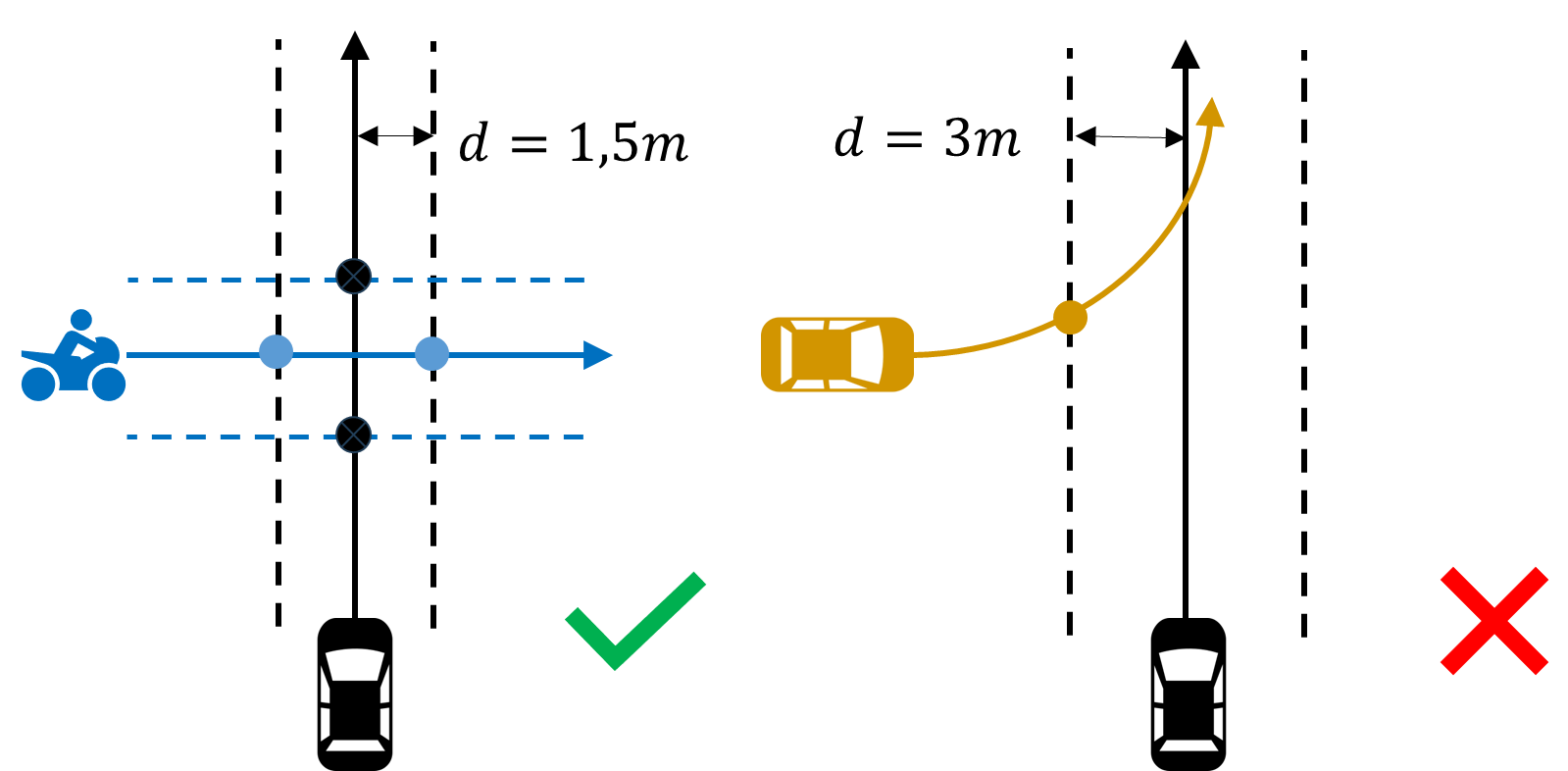}
\caption{Illustration of crossing trajectories identification.}
\label{fig: crossdemo}
\end{figure}

For applying the first rule, the method presented in Fig.\ref{fig: crossdemo} is used. For each of the two trajectories, two parallel buffer curves with distances to the trajectory of $d$ are added. One trajectory must intersect both parallel buffer curves of the other, to be considered as crossing. We set $d = \SI{3}{\meter}$ (approximately the width of a lane) for vehicles. For vulnerable agents such as pedestrians and cyclists, we set $d=\SI{1.5}{\meter}$ because they require smaller road space. This approach can effectively exclude undesired conflict types (merging and lane-changing).

For the second rule, we set a critical minimum distance of \SI{8}{\meter} between the centroids of two agents. \SI{8}{\meter} is approximately the minimum acceptable distance in urban environment \cite{jiao2023inferring}. A surrogate safety metric, Post-Encroachment-Time (PET) is also used. PET refers to the time difference between an encroaching road agent leaving the conflict area and another road agent entering the same area. According to the analysis of the relationship between crashes and PET in Peesapati et al. \cite{peesapati2013evaluation}, \SI{1.5}{\second} is a reasonable threshold to distinguish serious and normal conflicts in unprotected left-turn cases because it maximizes the correlation. However, serious conflicts are rare in practice and most serious conflicts are resolved before happening. Another concern is that ArgoAI AVs may drive more conservatively than human drivers. Therefore, this threshold is relaxed to \SI{5}{\second}. We choose \SI{5}{\second} because we want to preserve some space for further selection. Note that there are various approaches to calculating PET, for example, defining the conflict area as a reference. In this study, considering the diversity of intersection layouts and conflicting angles, we use the crossing point of the trajectories as the anchor (conflict point). Or one can say the PET here is equivalent to the Gap Time (GT). In practice, PET is smaller than GT. This is another reason why we relax the threshold. Consequently, we exclude those scenarios with PET $>\SI{5}{\second}$ \emph{and} a minimum distance $>\SI{8}{\meter}$.

The third rule introduces additional constraints. Firstly, at least one of the conflicting agents must traverse a distance greater than \SI{8}{\meter}. This excludes those cases influenced by great fluctuations caused by sensors, such as when both agents are situated at the boundary of the autonomous vehicle's perception range. Secondly, if the PET exceeds \SI{3}{\second}, then it is required that at least one of the conflicting agents exhibits a speed variation greater than \SI{3}{\meter\per\second} prior to reaching the conflict point. This assumption is rooted in the belief that for such long PETs if neither of the agents breaks or accelerates, there are no conflict resolution behaviours at all (but there might be a conflict).

After implementing these 3 rules, the chosen cases and conflicting agents are selected. The process of conflict resolution may involve considering surrounding road agents, even if their trajectories do not directly intersect with those of the conflicting agents. Therefore, it is necessary to include road agents surrounding the conflict point. This is achieved by drawing a circle with a radius of \SI{30}{\meter} around the conflict point, encompassing all agents with trajectories intersecting or fully contained within it. In Table.\ref{tab: CR}, we present the number of scenarios and the average number of surrounding agents for AV-involved and AV-free cases. Next, we will assess the quality of the raw data in Argoverse-2.



\begin{table}[h]
    \caption{Selected scenarios from Argoverse-2}
    \label{tab: CR}
    \begin{center}
    \begin{tabular}{l|cc}
    \toprule
    \textbf{Category} & \textbf{Nb. of scenarios}  & \textbf{Average Nb. of agents} \\ 
    \midrule
    AV-involved & 5337 & 20.5\\
    AV-free & 16094 & 21.7\\
    \bottomrule
    \end{tabular}
    \end{center}
\end{table}

\subsection{Raw data assessment}
\label{sec: raw data}

The quality of raw data must be carefully examined to make sure that further behavioural analysis is reliable. The motion of AVs and surrounding agents is obtained in different ways. For AVs, their positions and speeds can be directly reconstructed from the CAN Bus records. However, the motion of surrounding agents is usually processed from perception data, e.g. camera and LiDAR. Therefore, it is necessary to assess the quality of the trajectories separately for AVs and other agents. Argoverse-2 provides the location $(x,y)$, velocity $(v_x, v_y)$, and heading $\phi$ for each non-static agent. In this subsection, we mainly examine the consistency between positions and speeds.

\begin{figure}[h!]
\centering
\includegraphics[width=0.6\linewidth]{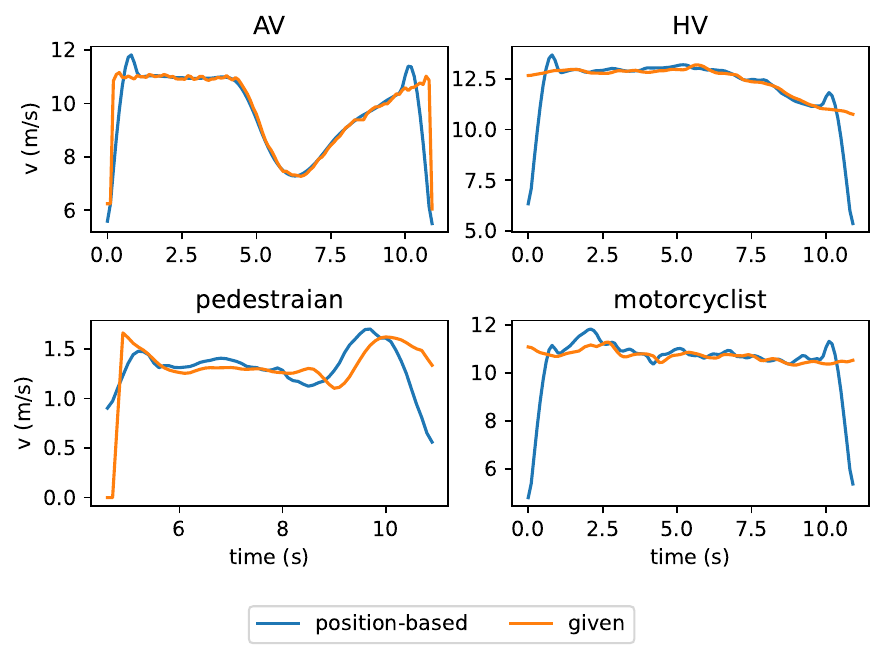}
\caption{An example: comparison of the given speed and the position-based speed for 4 different types of agents.}
\label{fig: trajexample}
\end{figure}

The speed derived from position series, the so-called \emph{position-based speed}, for example at moment $t$, is given by:
\begin{equation}
\begin{aligned}
    v_p &= \frac{s_{t-1, t} + s_{t, t+1}}{2\Delta t}\\
    s_{t-1, t} &= \sqrt{(x_t - x_{t-1})^2 + (y_t - y_{t-1})^2}
    \end{aligned}
\end{equation}
In Fig.\ref{fig: trajexample}, the position-based speed is compared with the given speed for 4 different types of agents in one of the selected scenarios. The comparison shows that the raw data in Argoverse-2 are not internally consistent. In fact, this inconsistency is observed for almost every non-static agent in Argoverse-2. We observe that the first and the last \SI{1.5}{\second} of the position series are not correctly processed. For every trajectory in Argoverse-2, position-based speed has a significantly sharp drop at the beginning and the end of each scenario. The drop magnitude is always around half of the nearest peak value. This type of error can be produced by using improper processing methods, such as fitting a polynomial with boundary constraints, using over-fitted wavelet denoising smoothers, or just using a pre-trained neural network. 

\begin{figure}[h!]
\centering
\includegraphics[width=0.33\linewidth]{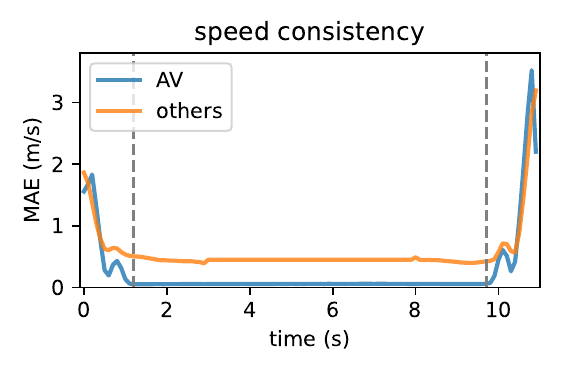}
\includegraphics[width=0.33\linewidth]{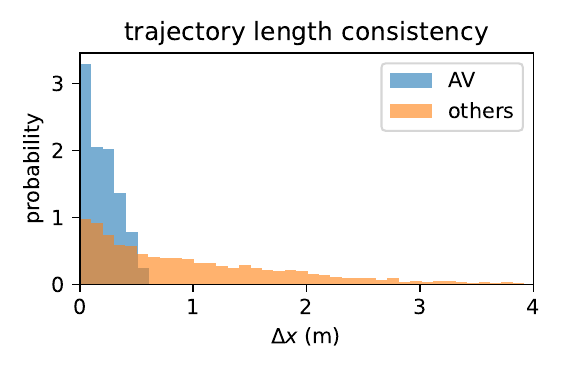}
\caption{The inconsistency between the given speed and the position-based speed (left) and the lengths of given trajectories and the trajectories reconstructed from the given speed (right). AVs and other agents are plotted separately.}
\label{fig: consistency}
\end{figure}

Fig.\ref{fig: consistency} quantifies the inconsistency of speed and trajectory length, respectively. The left figure shows that the first and the last \SI{1.5}{\second} of the trajectories have significantly higher MAE (between the given speed and the position-based speed). In the middle, AV's data are highly consistent but non-AV agents demonstrate higher inconsistency. The right plot of Fig.\ref{fig: consistency} presents the distribution of trajectory length inconsistency, which is the difference in travelling distances between the given trajectory and the trajectory that is constructed from the given speed by numerical integration. For AVs, the difference is always smaller than \SI{0.6}{\meter}. However, for non-AV agents, the difference is significantly larger and shows a fat-tailed distribution. This inconsistency is problematic for further analysis of driving behaviours.

The inconsistency brings a critical question: \emph{Speed or position, which one should we trust?} It is noticed that the given speed series are noisier than the position-based speed series, especially for AVs. For surrounding agents, some missing speed data was filled by 0 (this is also reported in Lyft level-5 raw data \cite{li2023large}). Compared with position-based speed, the given speed does not have strange drops but only the isolated outlier segments induced by 0-value padding (missing values). Therefore, we infer that \emph{the provided speed data is more reliable than positions.} This is the basic assumption for the following data enhancement procedure. 

\subsection{Data enhancement}

\begin{figure}[h!]
\centering
\includegraphics[width=0.5\linewidth]{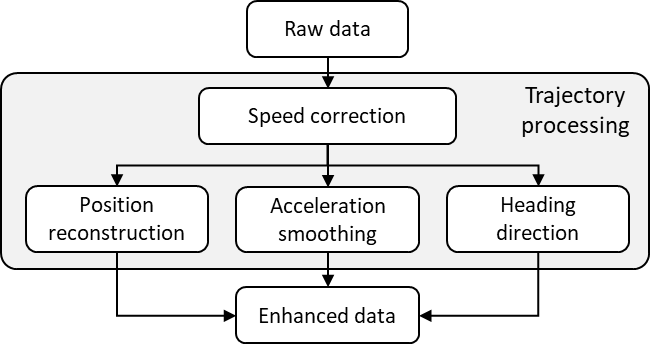}
\caption{Trajectory processing and enhancement.}
\label{fig: traj_process}
\end{figure}

This subsection describes how we correct and enhance the raw data. This encompasses the processing of both trajectory data and vectorized maps.
The flowchart of trajectory processing is presented in Fig.\ref{fig: traj_process}. AVs and non-AV agents share the same procedure but are processed differently after speed correction.

The process starts from the (assumed) more reliable speed data. The first step is speed outlier correction. At moment $t$, the speed $v(t)$ is an outlier if the acceleration is abnormally high or any speed in the neighbouring time interval is filled with 0:
\begin{equation}
\begin{aligned}
    &|\frac{dv(t)}{dt}| > \SI{10}{\meter\per\second\squared}\ \ \  \text{or}\\ 
    &\exists\ t' \in [t-\SI{0.3}{\second}, t+\SI{0.3}{\second}], \ \text{s.t.}\ \ v(t') = 0
    \end{aligned}
\end{equation}

We employ an empirical threshold of \SI{0.3}{\second} based on a visual inspection of the raw data. To rectify these outliers, we make the approximation that the jerk remains constant around the outliers, which means the speed is a cubic polynomial of time. The outliers are thus fitted from observations within the nearest \SI{1}{\second} window and subsequently utilized to substitute the raw values. Given that the majority of outliers consist of isolated points or short segments lasting less than \SI{0.3}{\second}, this approximation introduces minimal error. The corrected speed is denoted as $v'(t)$.

The second step is position reconstruction. For AVs, the first and the last \SI{1.5}{\second} of the position series are reconstructed from the corrected speed. For example, when reconstructing the last \SI{1.5}{\second}, we denote the initial x-position as $x_0$, and derive new x-positions after $T$ by:
\begin{equation}
    x(T) = x_0 + \int_0^T v'_x(t) dt
\end{equation}
The integration here can be numerically computed by Simpson's rule \cite{filon1930iii} and the same formula is used for y-positions as well. For non-AV agents, because their position and speed are inconsistent throughout the entire trajectory, we have to completely reconstruct the position data. We maintain the main trend of the given position but re-segment the trajectory based on minimising the change in corrected speed within each time interval (\SI{0.1}{\second}). First, the average speed of each time interval is computed. Next, the trajectory is re-segmented based on the polyline and the average speeds along time. These segment points are the reconstructed new position series. This method leads to naturally consistent speed and position data. However, not all lost information can be corrected in this way. For the following situations, we do not correct trajectories but preserve the raw data, even if the quality is poor:
\begin{itemize}
    \item The duration is shorter than \SI{5}{\second}. The unreliable percentage is too high so we cannot correct the data.
    \item The length of the trajectory is shorter than \SI{8}{\meter}. The perception fluctuation is too large. This only applies to non-conflicting background agents.
    \item The trajectory length inconsistency is larger than \SI{2}{\meter}, which means the error is too large.
\end{itemize}

Next, the corrected speed is smoothed, from which we derive acceleration and heading direction. For AVs, considering that the speed is noisy, we use the wavelet denoise method \cite{pan1999two} to smooth the speed and derive acceleration by computing gradient. Here we use the wavelet denoising in \texttt{skimage} python package. The noise is set to $\sigma = \SI{0.5}{\meter\per\second}$ (by trial-and-error). The wavelet mode is the Daubechies family with 6 vanishing moments (`\texttt{db6}'). The soft threshold method is used and the maximum wavelet decomposition level is 3. For non-AV agents, the corrected speed is smooth enough already, so we directly compute the gradient as acceleration. The heading is defined as the tangent direction of the trajectory. Because the bounding boxes of vehicles are not given in the raw data, we cannot further process the yaw direction.

\begin{figure}[h!]
\centering
\includegraphics[width=0.6\linewidth]{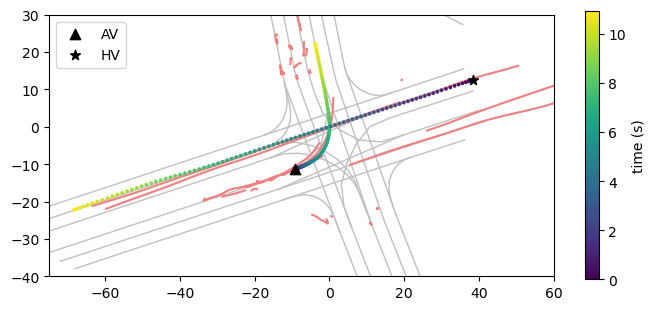}
\includegraphics[width=0.66\linewidth]{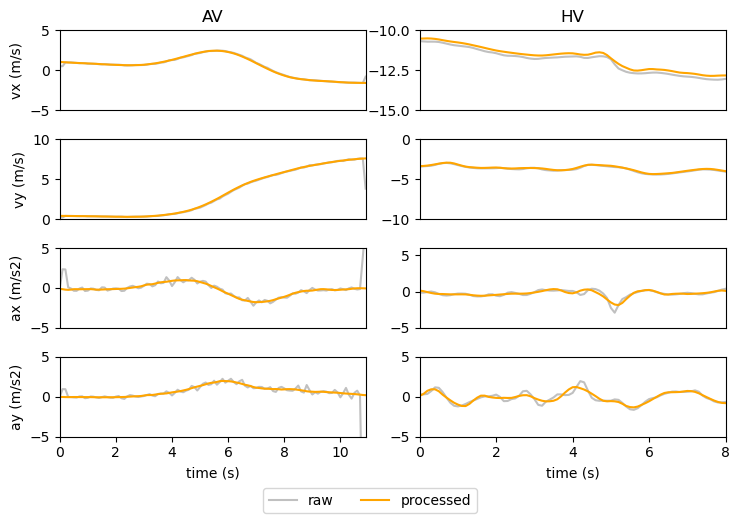}
\caption{Example of comparison between raw and processed motion data.}
\label{fig: example}
\end{figure}

As an example, Fig.\ref{fig: example} compares the raw and the processed motion data for an AV and an HV in the case of an unprotected left turn. The outliers of speed at both ends are removed and the speed and acceleration are smoother after processing.

The Argoverse-2 data also provides vectorized lane geometries along the driving direction. The average length of a lane segment is around \SI{15}{\meter}, which means that there are too many segments in each scenario. To save storage space and preserve road connectivity and traffic rules, we combine and merge the lane segments by the following rules:
\begin{itemize}
    \item Segments are merged together if one is the only downstream segment of the other one.
    \item All combined lanes start and end at the boundary of the maps, a diverging point, or a merging point.
    \item All combined lanes are re-segmented into a series of 20 tail-to-head vectors along the driving direction.
\end{itemize}

\begin{figure}[h!]
\centering
\includegraphics[width=0.3\linewidth]{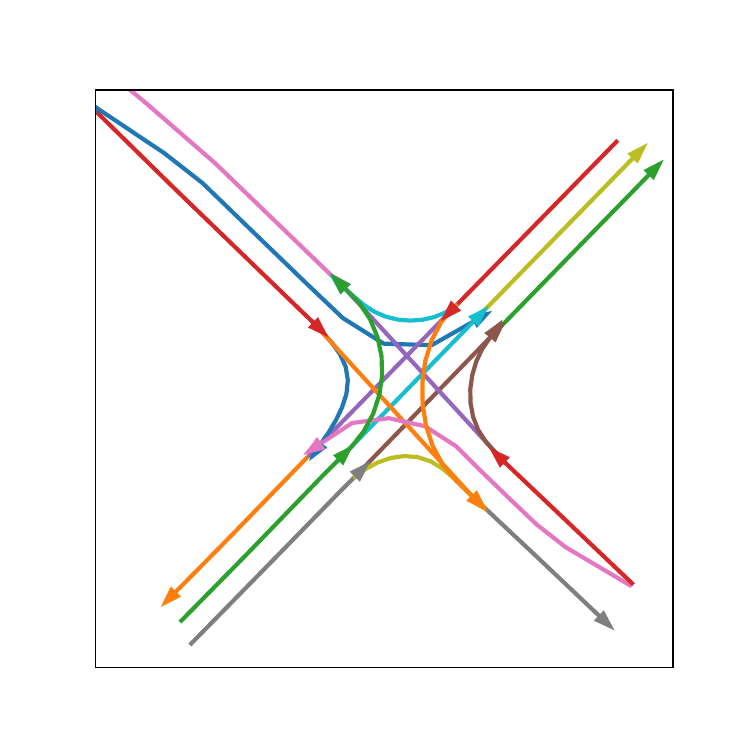}
\includegraphics[width=0.3\linewidth]{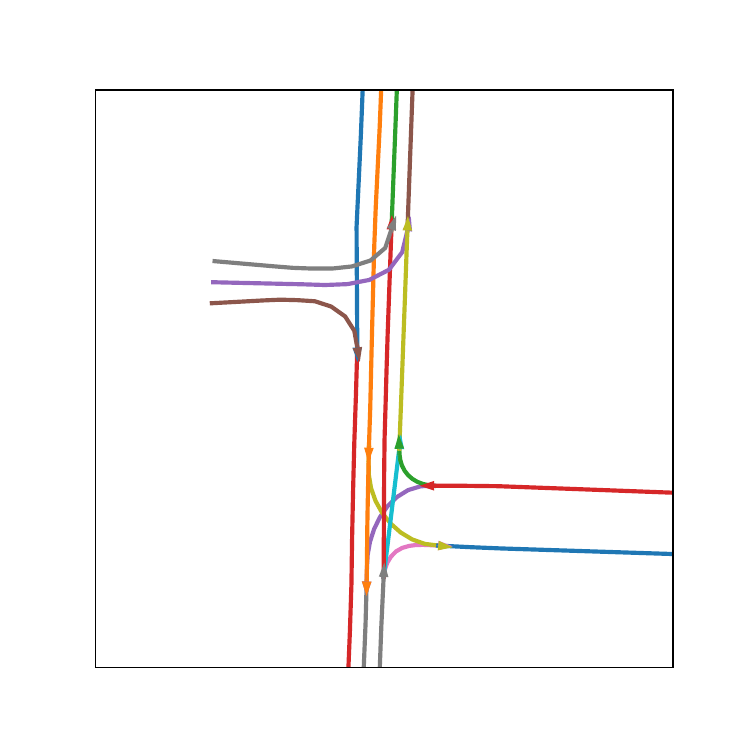}
\caption{Two examples of processed vectorized maps.}
\label{fig: mapdemo}
\end{figure}

Using these rules, road maps can be stored in a compact array and the adjacency matrix of all combined lanes can be easily constructed through indexing the start/end positions. Two examples of the processed maps are shown in Fig.\ref{fig: mapdemo}.  Different colours represent different segments of lanes. The arrow at the end of each segment indicates the driving direction.

The introduced data selection and enhancement procedure given here results in an enhanced dataset. The dataset is publicly available on the 4.TU data platform \footnote{https://github.com/RomainLITUD/conflict\_resolution\_dataset}.

\section{Dataset assessment}
\label{sec: assessment}

With the dataset obtained, now we focus on assessing the quality of trajectories and the diversity of conflict resolution regimes. This step is crucial in ensuring that the data is reliable and rich for further analysis of safety and efficiency impacts. Also, for user-friendliness, classifying and labelling the regimes allow researchers to select the focused type of conflicts.

\subsection{Anomaly analysis}

In addition to speed inconsistency introduced in Section \ref{sec: raw data}, we consider 3 constraints on acceleration $a$ and jerk $j$ proposed by Punzo et al. \cite{punzo2011assessment} to assess the quality of vehicle trajectories:
\begin{itemize}
    \item $a\in [-8, 5]\ \SI[per-mode=repeated-symbol]{}{\meter\per\second\squared}\label{eq: range}$
    \item $j\in [-15, 15]\ \SI[per-mode=repeated-symbol]{}{\meter\per\cubic\second}$
    \item The jerk's sign cannot reverse more than once in \SI{1}{\second} (\textit{Jerk Sign Inversion}, JSI).
\end{itemize}

Violating these constraints on vehicle kinematics is considered an ``anomaly''. Table.\ref{tab: anomaly} presents the speed inconsistency ($\Delta v$) and the time percentages of detected anomalies of acceleration, jerk, and JSI for vehicles. Only the conflicting vehicles are included in this table. The enhanced data is compared with the raw data. The result clearly shows that the quality of the enhanced data is indeed significantly improved compared to the raw data in Argoverse-2. The consistency increases and the time percentage of anomaly, especially JSI anomaly, significantly decreases. Pedestrians' and cyclists' motion has higher flexibility so we do not assess their data quality here.

\begin{table}[h]
    \caption{Anomaly assessment}
    \label{tab: anomaly}
    \begin{center}
    \begin{tabular}{l|cccc}
    \toprule
    \textbf{Dataset} & \textbf{$\Delta v$} (\SI{}{\meter\per\second})  & \textbf{acc} \% & \textbf{jerk} \% & \textbf{JSI} \%\\ 
    \midrule
    AV-enhanced & 0.01 & 0.01 & 0.01 & 0.3\\
    HV-enhanced & 0.30 & 0.01 & 0.03 & 0.9\\
    \midrule
    AV-raw & 0.32 & 0.21 & 0.23 & 83.6\\
    HV-raw & 3.30 & 0.14 & 0.19 & 13.9\\
    \bottomrule
    \end{tabular}
    \end{center}
\end{table}

\subsection{Regime classification}

\begin{figure}[h!]
\centering
\includegraphics[width=0.5\linewidth]{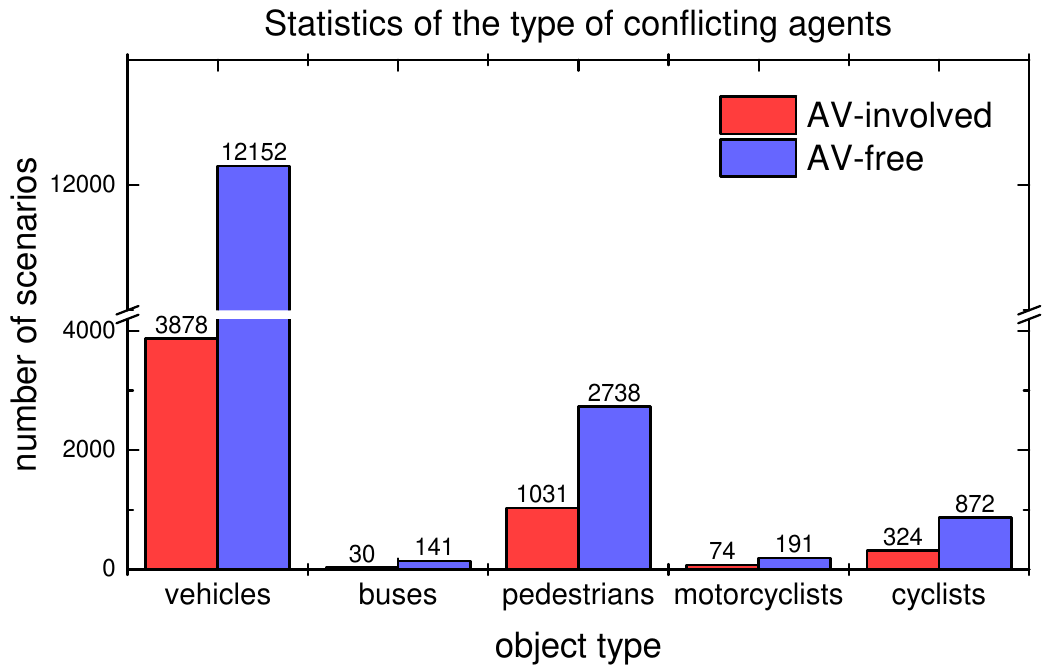}
\caption{Distributions of different types of conflicting agents}
\label{fig: conflict obj}
\end{figure}

The diversity of selected scenarios is evaluated by the type of conflicted agent (with an AV or without AV, conflict with a vehicle or a pedestrian, etc.) and also the conflict regimes (based on incoming and intended directions). In Fig.\ref{fig: conflict obj}, the statistics of the types of the other conflicting agents for AV (red bars)  and HV (blue) are presented. The results show that the dataset comprises conflicts between diverse agent types, e.g. pedestrians and cyclists. Especially, vehicle-vehicle and vehicle-pedestrian conflicts compose the majority of the dataset.

\begin{figure}[h!]
\centering
\includegraphics[width=0.5\linewidth]{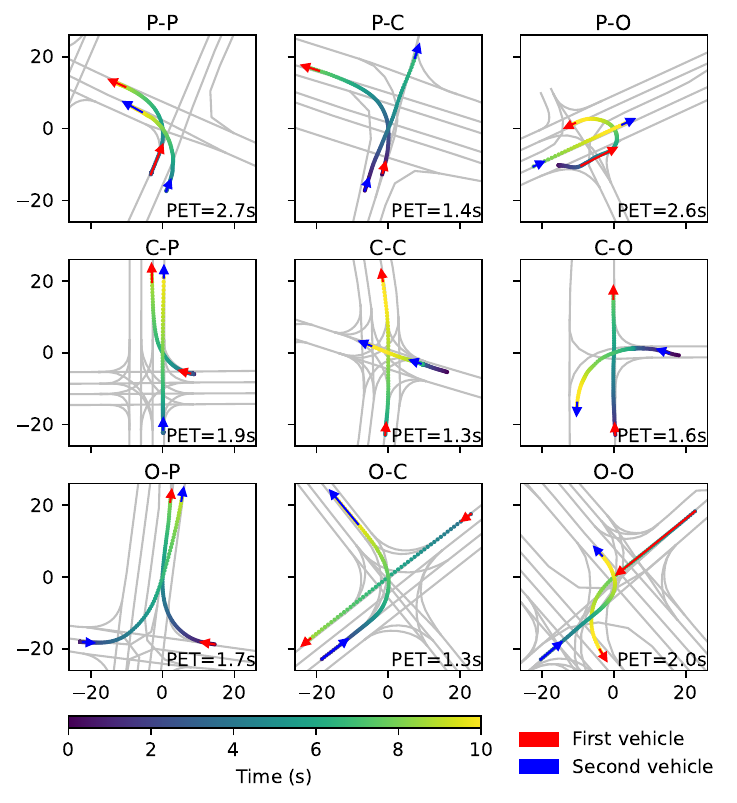}
\caption{Nine combinations of conflict regimes}
\label{fig: regime}
\end{figure}

Considering the functionality of redistributing traffic flow for intersections, conflict regimes in the scenarios are classified based on the relative movement of the two vehicles, i.e., whether they run parallel (\textbf{P}), cross (\textbf{C}), or run opposite (\textbf{O}) to each other before and after reaching the conflict point. Consequently, there are 9 combinations as shown in Fig. \ref{fig: regime}. Further, we differentiate these combinations by considering whether the second-passing vehicle was from the left or the right of the first-passing vehicle. For simplification, we use a notation of two letters pointed by an arrow to represent these cases. For example, \textbf{P}$\rightarrow$\textbf{C} indicates that the two vehicles were running parallel before the conflict point and crossing afterwards, and the second-passing vehicle moved from the left to the right of the first-passing vehicle. 

\begin{figure}[h!]
\centering
\includegraphics[width=0.32\linewidth]{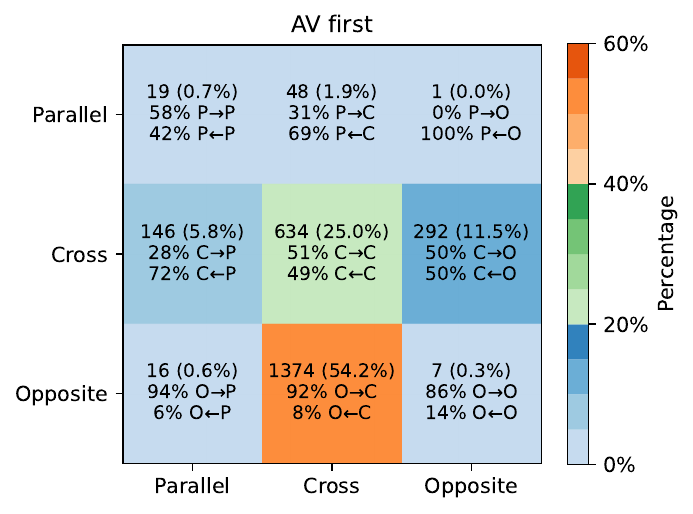}
\includegraphics[width=0.32\linewidth]{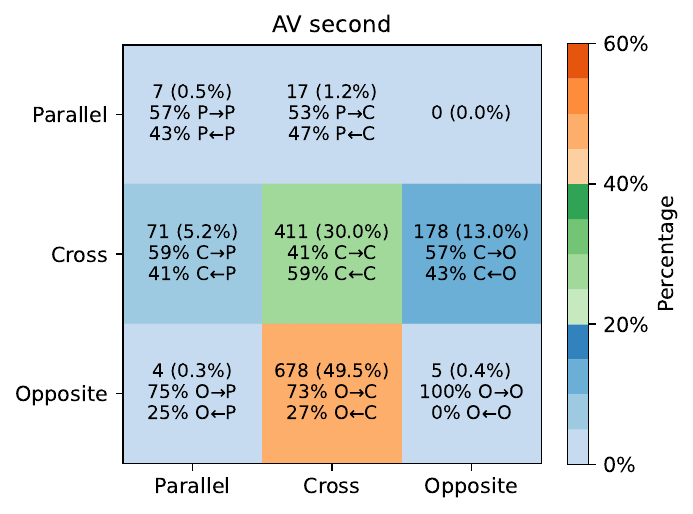}
\includegraphics[width=0.32\linewidth]{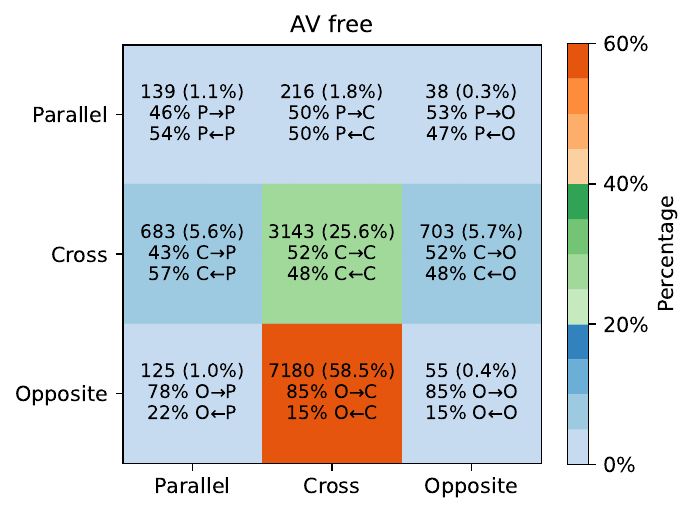}
\caption{Statistics of conflict regimes for different scenarios.}
\label{fig: regime_stat}
\end{figure}

We further consider 3 types of scenarios to observe the distribution of these regimes. (1) An AV passing the conflict point before a human agent (\emph{AV-first}); (2) an AV passing the conflict point after a human agent (\emph{AV-second}); (3) the conflict happens between an HV and a human agent (\emph{AV-free} or \emph{HV}). Fig.\ref{fig: regime_stat} presents the statistics of vehicle-vehicle conflicts of the 3 types of scenarios, showing similar regime distributions across them. Around and more than 80\% of conflicts fall into  \textbf{O}-\textbf{C} and \textbf{C}-\textbf{C} types. Among all the regimes, \textbf{O}$\rightarrow$\textbf{C} is the most common type. Most unprotected left turns belong to this category. The result shows that the dataset contains diverse and balanced conflict regimes for both AV-involved and AV-free scenarios. We also provide the initial and ending direction vectors of each agent in the enhanced dataset so users can re-categorize the conflict regimes based on their needs.

In summary, the enhanced dataset contains highly consistent and smooth trajectory data and diverse and balanced conflict resolution regimes. We also label the conflict regimes for user-friendliness. This dataset is suitable for comparing the conflict resolution behaviours in AV-involved and AV-free scenarios. In the next section, we will conduct a preliminary analysis of the safety and efficiency impacts of AVs using the enhanced dataset.

\section{Impact assessment}
\label{sec: analysis}

In this section, we use the obtained enhanced dataset to assess and compare the safety and efficiency performances of AV-involved and AV-free conflict resolution cases. We especially focus on studying whether human road users react differently to the AV in conflict resolution.

\subsection{Safety assessment}

Wang et al. \cite{wang2021review} systematically reviewed the surrogate safety measures (SSM) and their applications on connected and automated vehicles. According to that review paper, there are 3 categories of SSM, namely time-based, deceleration-based, and energy-based measures. Energy-based measures are mainly used for assessing collision severity, which is not the focus of this study. Among the other time-based and acceleration-based SSMs, only PET, Gap Time (GT), and Proportion of Stopping Distance (PSD) are applicable for crossing/angle conflicts. As explained before, due to the missing vehicle size data, PET and GT are regarded as equivalent in this study. Next, we assess the safety impacts of AVs in vehicle-vehicle and vehicle-pedestrian interactions by analyzing PET and PSD. In principle, the assessment depends on the specific context, e.g. conflicting angles. Nevertheless, the previous section showed that the AV-involved and AV-free cases are well-balanced in terms of conflict regimes, so we omit the regimes but focus on three major factors, conflict types (with vehicle or pedestrian), AV/HV, and passing order (who goes first).

\begin{figure}[h!]
\centering
\includegraphics[width=0.26\linewidth]{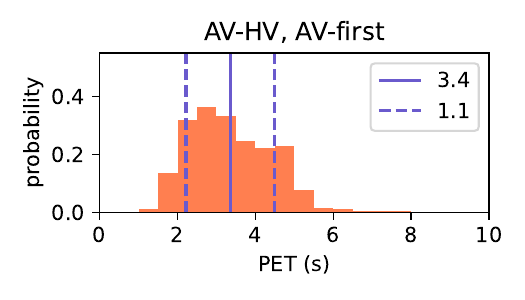}
\includegraphics[width=0.26\linewidth]{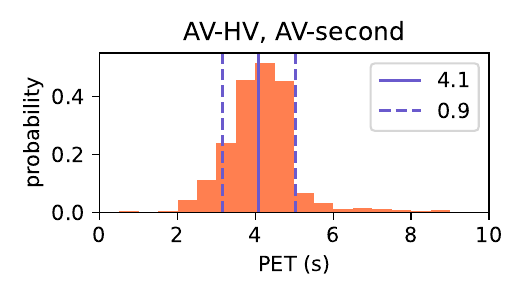}
\includegraphics[width=0.26\linewidth]{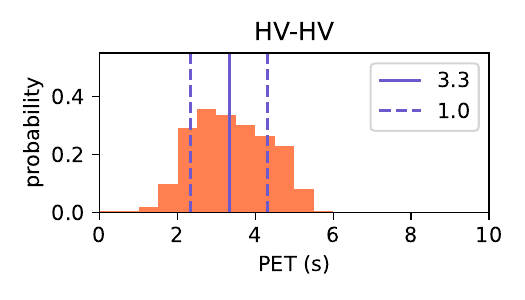}
\vfill
\includegraphics[width=0.26\linewidth]{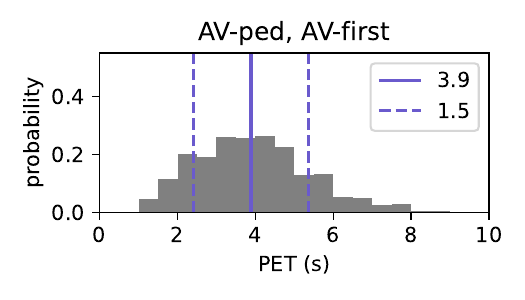}
\includegraphics[width=0.26\linewidth]{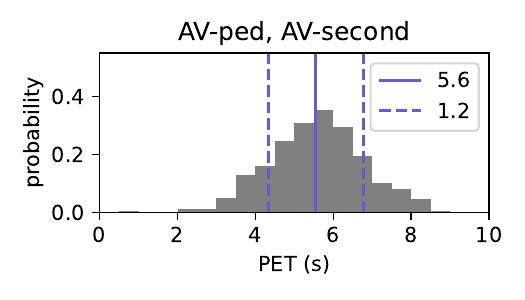}
\vfill
\includegraphics[width=0.26\linewidth]{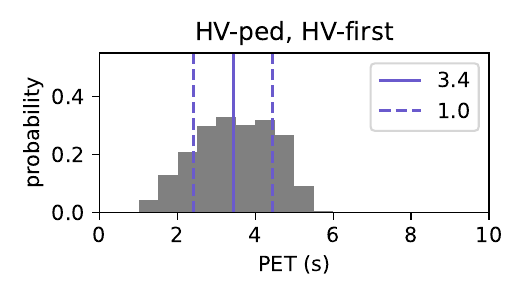}
\includegraphics[width=0.26\linewidth]{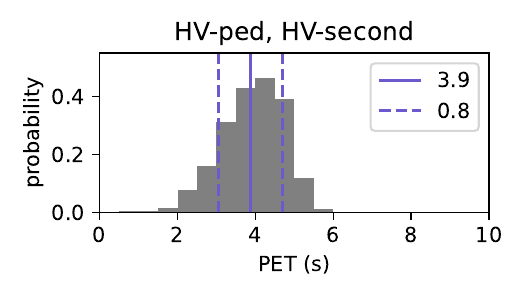}
\caption{Distributions of PETs for different scenarios of conflict resolution. The blue solid lines mark the mean values ($\mu$) and the dotted lines mark the standard variance widths ($\mu\pm \sigma$).}
\label{fig: pet_dist}
\end{figure}

The definition of PET was introduced in Section \ref{Sec: Processing}. We compare PET distributions for 3 scenarios of conflict resolution, i.e., AV-first, AV-second, and AV-free (HV-HV). The results are presented in Fig.\ref{fig: pet_dist}. For vehicle-vehicle interactions, we have two major observations from the top 3 figures in Fig.\ref{fig: pet_dist}. First, AV-first cases have almost the same distribution (close mean value and standard variance) as AV-free scenarios, but AV-second cases have significantly higher average PET (\SI{4.1}{\second}). Second, AV-first and AV-free scenarios contain more low-PET cases (< \SI{2}{\second}), but the PETs in the AV-second scenarios are always larger than \SI{2}{\second}. Considering that PET is mainly determined by the second-passing agent, the observation implies that the AV in Argoverse-2 is more conservative (and possibly safer) than HVs when interacting with vehicles. As for vehicle-pedestrian interactions, whether an AV is involved shows a significant difference. Both AV-first and AV-second cases have higher mean values and larger variance of PET than HV-involved counterparts. The difference between AV-second and HV-second scenarios in the vehicle-pedestrian group is more significant than in the vehicle-vehicle group. This result indicates that the AVs in Argoverse-2 behave even more cautiously when interacting with pedestrians. Furthermore, compared to HV-first, the pedestrians in AV-first scenarios show a long-tail PET distribution at the high-value end.  This observation implies that some pedestrians become more conservative when facing an AV than when facing an HV (keeping a longer distance or moving slower). So the overall effect is pedestrians behave more diversely when conflicting with AVs than with AVs.

PSD is defined as the ratio between the remaining distance to the potential collision point and the minimum acceptable stopping distance. Its formula is:
\begin{equation}
    \text{PSD} = \frac{d_t}{v^2_t/(2|a_{max}|)}
\end{equation}
In the initial study of Allen et al. \cite{allen1978analysis}, $d_t$ is the distance between two conflicting vehicles, $a_{max}$ is the maximum acceptable deceleration, and $v_t$ is the operation speed of the second-passing vehicle. At intersections, $d_t$ is defined as the longitudinal distance (along the lanes or the trajectory) to the conflict point. $a_{max}$ is set as \SI{-3.35}{\meter\per\second\squared} \cite{astarita2012new} for all vehicles. For each case, we compute how PSD changes in time until the following agent passes the conflict point, at which point the minimum value is selected. The smaller the PSD is, the more risky the manoeuvre is. Compared to PET, PSD inclines to measure the decelerating behaviour of the second-passing agent \emph{before} the other agent passes the conflict point. Pedestrians are not subject to an acceptable deceleration. Therefore, in vehicle-pedestrian interactions, we only consider the PSD of vehicles when pedestrians go first. 

\begin{figure}[h!]
\centering
\includegraphics[width=0.3\linewidth]{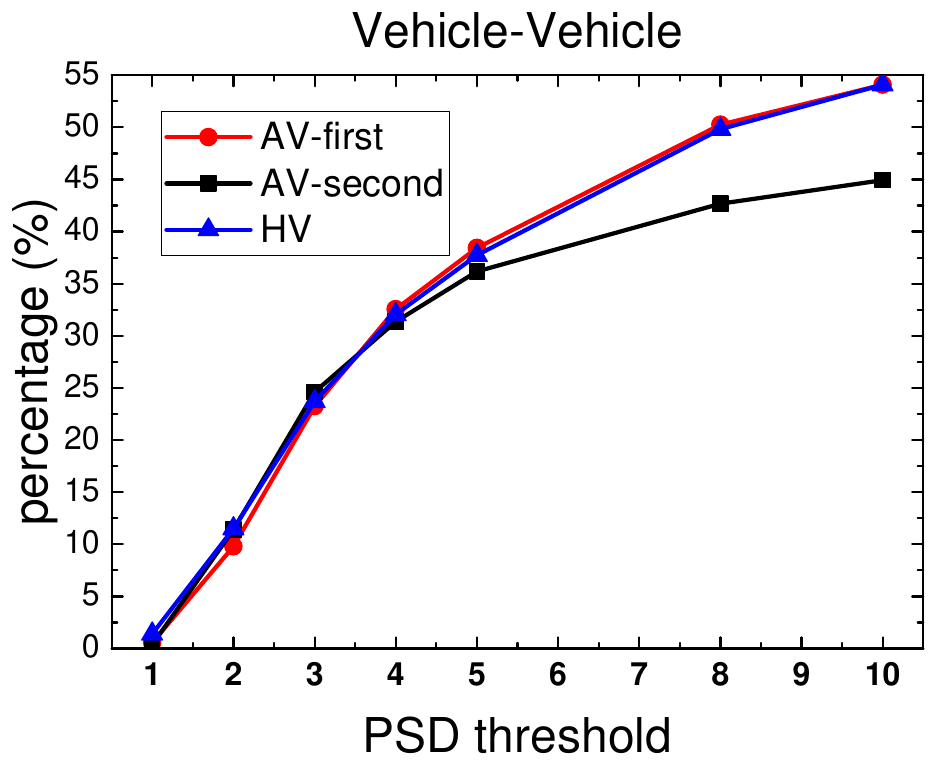}
\includegraphics[width=0.3\linewidth]{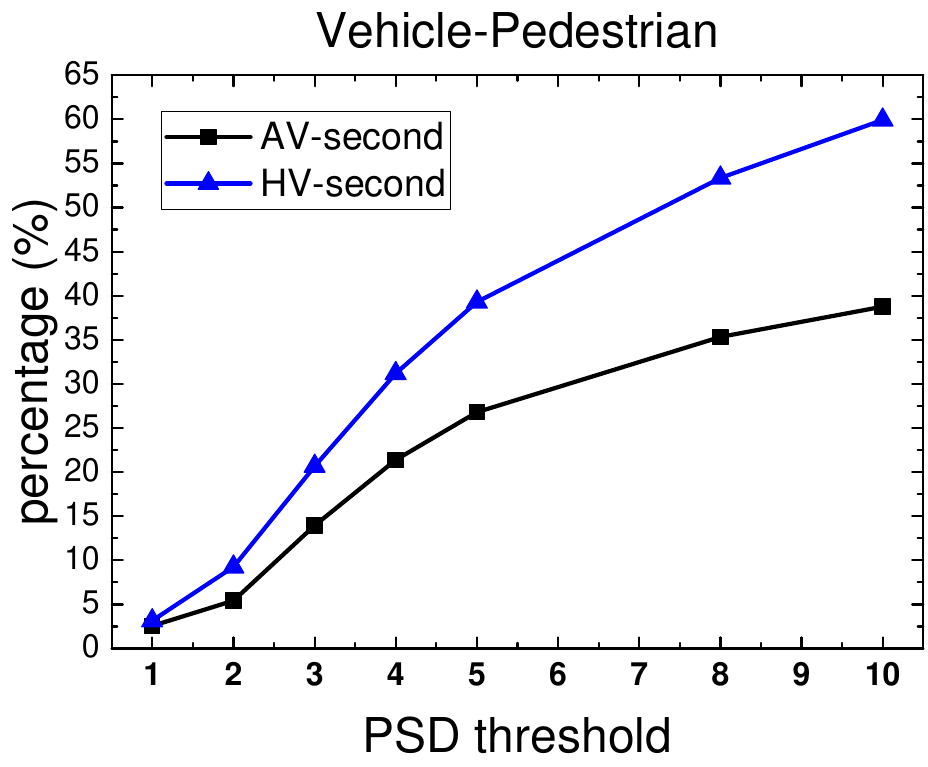}
\caption{PSD for different types of scenarios.}
\label{fig: psd}
\end{figure}

\begin{figure}[h!]
\centering
\includegraphics[width=0.25\linewidth]{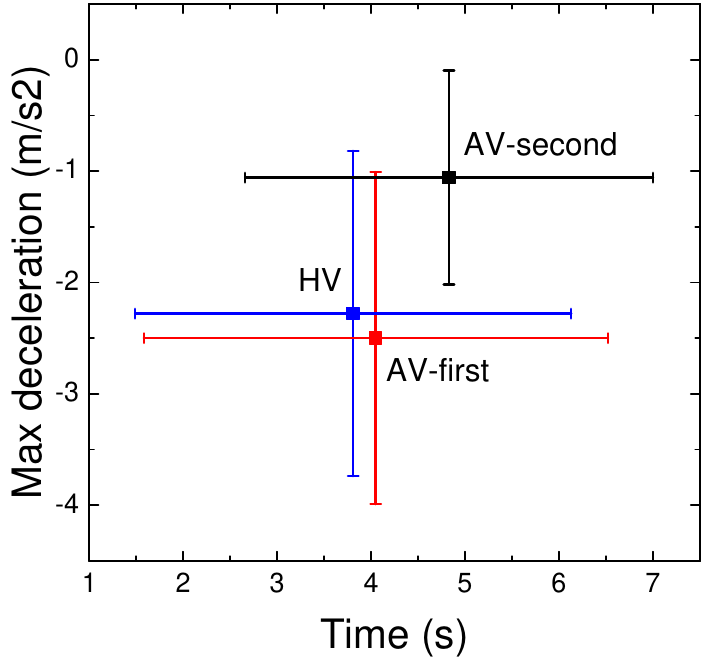}
\caption{Mean value and standard deviation of both the maximum deceleration and the time gap between passing the conflict point and reaching the maximum deceleration.}
\label{fig: time-dece}
\end{figure}

The plots in Fig.\ref{fig: psd} present how many conflicts in percentage have PSD smaller than different threshold values. For vehicle-vehicle interactions, the three scenarios have almost the same percentages when the PSD thresholds are lower than 4. However, PSD cannot completely reflect the differences in driving behaviours. Fig.\ref{fig: time-dece} further shows the mean and the standard deviation of the maximum deceleration before reaching the conflict point, and the time interval between the moment of reaching the maximum deceleration and the moment of passing the conflict point for the second-passing vehicle. We see that, although the 3 scenarios have similar PSD curves, AVs tend to brake earlier and more softly than HVs. As for HVs, similar to PET, we do not observe significant differences when interacting with an AV and an HV. HV and AV-first have close braking properties. For vehicle-pedestrian interactions, the percentages of HV-second cases are consistently higher than AV-second cases, which also implies that the AVs in Argoverse-2 drive more conservatively than HVs when interacting with pedestrians.

The safety assessment result is summarized as follows:
\begin{itemize}
    \item When AV passes first, HVs do not react differently regarding PET, PSD, and deceleration behaviours, compared to the situation where other HVs pass first. But, compared to HVs, when an AV pass after an HV, the AV drives more conservatively and thus potentially safer.
    \item Compared to vehicles, pedestrians show more significant behavioural adjustment to the presence of AVs. The difference is reflected in the larger average mean value (more cautious) and the wider range (more diverse reactions of pedestrians) of PET.
\end{itemize}

\subsection{Efficiency assessment: Minimum Recurrent Clearance Time}

Evaluating the effects of AVs on traffic flow at intersections holds equal significance alongside safety, as prioritizing safety should not come at the expense of diminished efficiency. A conflict resolution process can be divided into two stages, before and after one agent passes the conflict point (for convenience, we call them the \emph{pre-conflict} stage and \emph{post-conflict} stage). In this subsection, we propose a novel measure, the so-called \emph{minimum recurrent clearance time} (MRCT), to measure conflict resolution efficiency. The basic idea of MRCT is, assuming that the same interaction process repeatedly happens at the same location, what is the minimum required time to clear the conflict area for the next interaction? The measure is based on the following assumptions:
\begin{itemize}
    \item[A-1] Vehicles from two conflicting streams pass the conflict point alternately.
    \item[A-2] Each interaction between two conflicting vehicles is not influenced by the previous interaction.
    \item[A-3] The same conflict resolution process happens repeatedly in a uniform time interval around the conflict point.
\end{itemize}

Based on the 3 assumptions, we can derive the mathematical form of MRCT. For convenience, we use subscripts 1 and 2 to represent the 1st and 2nd passing vehicles in one conflicting vehicle pair, respectively. We use superscript $'$ to represent the current pair of vehicles and without $'$ means the previous pair of vehicles. The 4 vehicles in the adjacent two interactions are denoted as V1, V2, V1', and V2'.The meaning of the used notations are listed in Table.\ref{tab: notations}.

\begin{table}[h]
    \caption{List of notations used for computing MRCT}
    \label{tab: notations}
    \begin{center}
    \begin{tabular}{l|c}
    \toprule
    \textbf{Notation} & \textbf{Meaning}\\ 
    \midrule
    $t$ & time \\
    $t_i$ & moment of the vehicle $i$ passing the conflict point\\
    $s$ & curvilinear distance to the conflict point, negative when approaching the conflict point\\
    $v$ & speed\\
    $\Delta t^*$ & MRCT\\
    $d_g(v)$ & critical gap acceptance\\
    $d_h(v)$ & critical distance headway\\
    \bottomrule
    \end{tabular}
    \end{center}
\end{table}

Firstly, assumption A-1 implies that V1' and V2' are subjected to the trajectories of V1 and V2. The relationship between the vehicles on the same stream is car-following so we define a critical distance headway $d_h(v)$. From the perspective of one vehicle, it is also subjected to the vehicles from another stream. So, we also define a critical gap acceptance $d_g(v)$. If the distance headway and the gap are both longer than the predefined critical values, we say that the current interaction is not influenced by the previous interaction (assumption A-2). Further, according to assumption A-3, if the time interval between two consecutive interaction processes is $\Delta t$, then the distance and speed of V1' and V2' can be expressed by the motion of V1 and V2 as follows:
\begin{equation}
\begin{aligned}
    s_i'(t) &= s_i(t-\Delta t)\\
    v_i'(t) &= v_i(t-\Delta t)
\end{aligned}
\label{eq: alternative}
\end{equation}

For vehicle V1', the following conditions must be met:
\begin{equation}
\begin{aligned}
    s_1(t) - s_1'(t) & \geq d_h(v_1'(t)), \ \ \ \forall t \leq t_1\\
    s_1'(t_2) & \geq d_g(v_1'(t_2))
\end{aligned}
\end{equation}
For vehicle V2', we have the constraints:
\begin{equation}
    s_2(t) - s_2'(t) \geq d_h(v_2'(t)), \ \ \ \forall t \leq t_2
\end{equation}
By using Eq.\ref{eq: alternative}, we can write:
\begin{equation}
\begin{aligned}
    s_1(t) - s_1(t-\Delta t) & \geq d(v_1(t-\Delta t)), \ \ \ \forall t \leq t_1\\
    s_1(t_2-\Delta t) & \geq d(v_1(t_2-\Delta t))\\
    s_2(t) - s_2(t-\Delta t) & \geq d(v_2(t-\Delta t)), \ \ \ \forall t \leq t_2
\end{aligned}
\label{eq: constraints}
\end{equation}

The minimum value of $\Delta t$ that satisfies Eq.\eqref{eq: constraints} gives MRCT:
\begin{equation}
    \text{MRCT} = \Delta t^* = \min \Delta t\ \  \text{s.t. Eq.\eqref{eq: constraints} hold}
    \label{eq: opt}
\end{equation}

\begin{figure}[h!]
\centering
\includegraphics[width=0.5\linewidth]{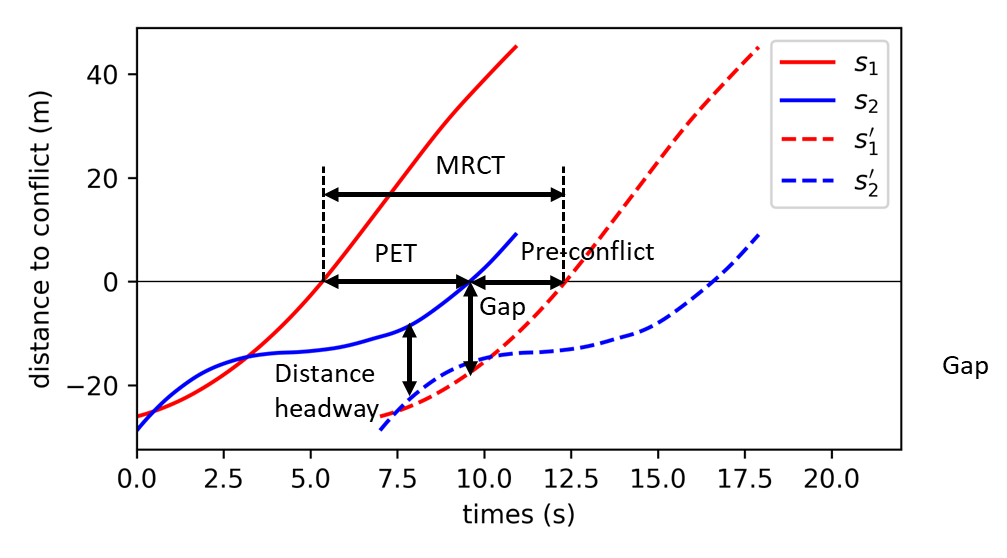}
\caption{Illustration of concepts for computing MRCT.}
\label{fig: mrct-demo}
\end{figure}

These concepts are illustrated in Fig.\ref{fig: mrct-demo} for a more intuitive understanding. In this way, we convert the efficiency evaluation problem into an optimization problem. If there is no conflict at all, this model degrades to the basic car-following model with a critical distance headway. It is straightforward that MRCT $\geq$ PET. If PET is interpreted as the post-conflict passing time that is mainly decided by the second-passing vehicle, MRCT$-$PET can be understood as the negotiation time of the first vehicle and thus the conflict resolution efficiency. We call it \emph{pre-conflict duration}. If the negotiation takes too long (for example, both vehicles decelerate and hesitate before determining the passing order), or if the first-passing vehicle does not accelerate to pass the conflict point faster, the pre-conflict duration will be longer. 

The form of $d_h(v)$ and $d_g(v)$ is calibrated from field test data or simulators. Here we use an empirical and simplified linear relationship adopted from a car-following model \cite{bose1999analysis}:
\begin{equation}
\begin{aligned}
    d_h(v) &= 2\cdot v + 8\\
    d_g(v) &= \max (2\cdot v, 8)
\end{aligned}
\end{equation}

Note that problem \eqref{eq: opt} does not have a solution if a segment of the trajectories is missing or the pre-conflict trajectory of the second-passing vehicle is too short. We omit these invalid cases. 

\begin{figure}[h!]
\centering
\includegraphics[width=0.33\linewidth]{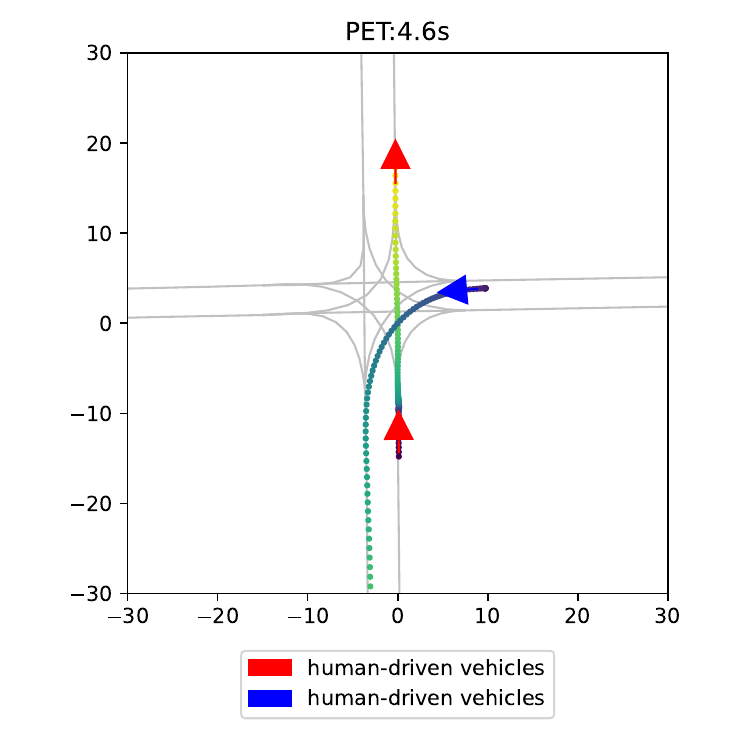}
\includegraphics[width=0.33\linewidth]{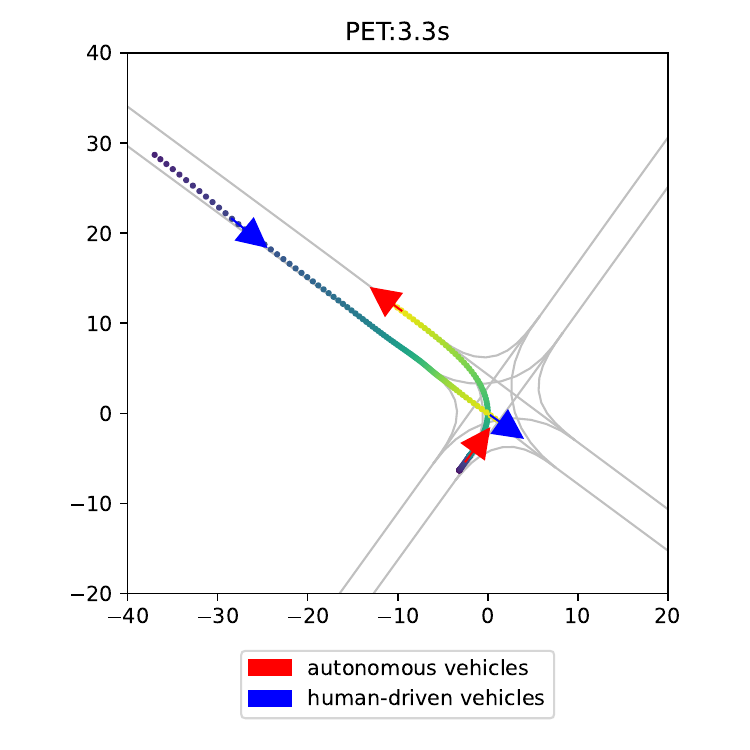}
\includegraphics[width=0.33\linewidth]{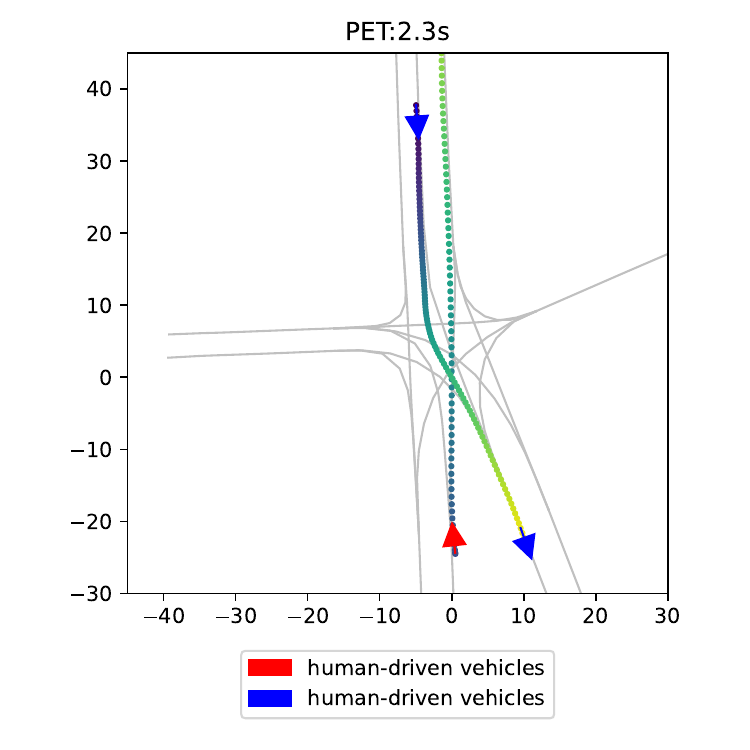}
\vfill
\includegraphics[width=0.33\linewidth]{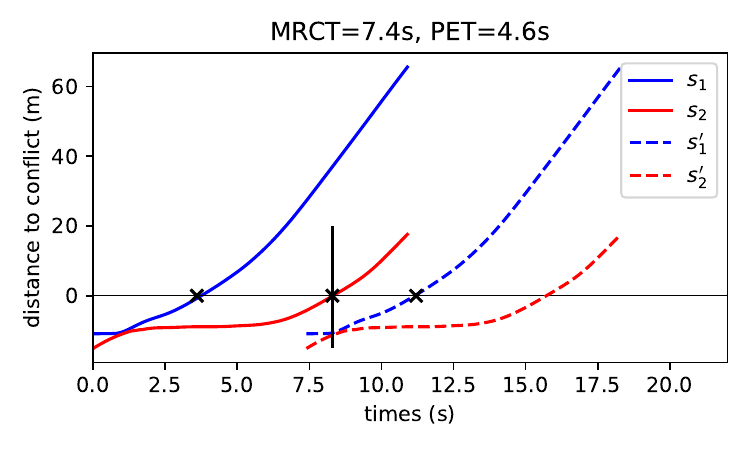}
\includegraphics[width=0.33\linewidth]{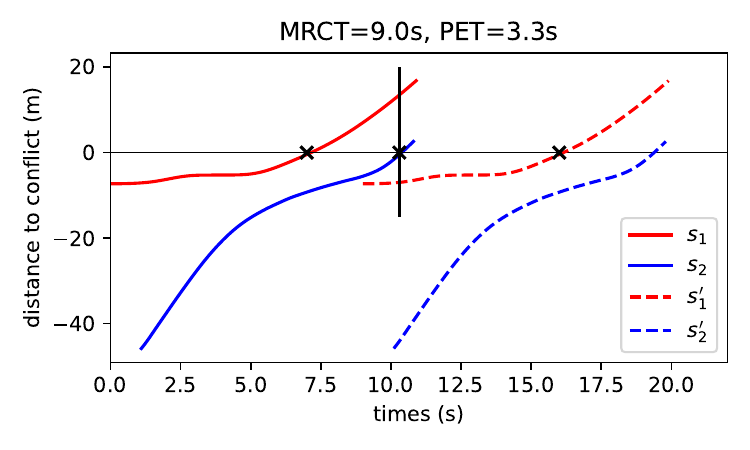}
\includegraphics[width=0.33\linewidth]{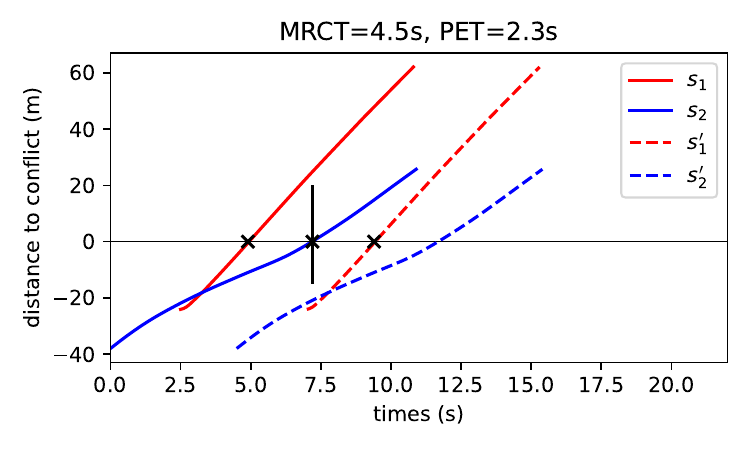}
\caption{Three examples of conflict resolution. The black lines separate PET from pre-conflict duration.}
\label{fig: mrct-example}
\end{figure}

To demonstrate the validity of the MRCT, we show 3 examples in Fig.\ref{fig: mrct-example}. The first example has a long PET, but a shorter pre-conflict duration. The passing order is determined quickly because we see that the red vehicle brakes immediately without hesitation and the blue vehicle accelerates to pass the conflicting point faster. However, the red HV stands still and waits for several seconds, which causes a long PET. The second example has a long pre-conflict duration (\SI{5.7}{\second}) and a relatively shorter PET (\SI{3.3}{\second}). From the trajectory profiles, we see that there is a long negotiation process. Both vehicles decelerate and finally, the left-turning AV passes first instead of waiting for the straight-ahead HV (which has the priority to pass according to traffic rules). The third example is a highly efficient case of human interaction. Both PET and pre-conflict duration are short. Only the blue vehicle slightly decelerates (due to turning or conflicts) and the two vehicles pass the conflict point quickly. Therefore, MRCT can effectively capture the efficiency of conflict resolution based on experience and domain knowledge.

\begin{figure}[h!]
\centering
\includegraphics[width=0.4\linewidth]{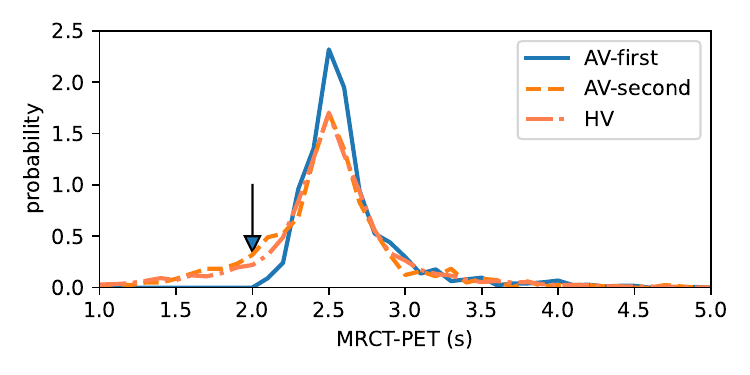}
\includegraphics[width=0.4\linewidth]{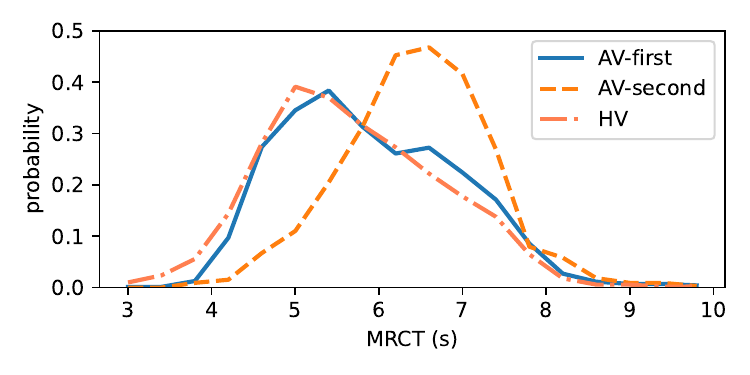}
\includegraphics[width=0.4\linewidth]{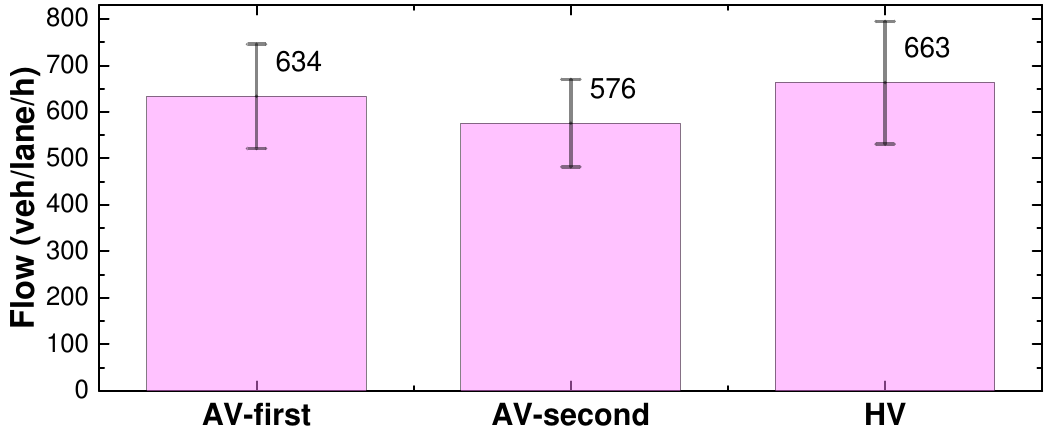}
\caption{The upper two figures show the distributions of pre-conflict duration and MRCTs for different types of conflict resolution scenarios. The bottom bar chart presents the mean and the variance of the estimated average flow for different types of conflict resolution.}
\label{fig: efficiency}
\end{figure}

The upper 2 plots in Fig.\ref{fig: efficiency} present the distributions of pre-conflict duration and MRCT for different scenarios. For all the scenarios, the pre-conflict duration concentrates around \SI{2.5}{\second}. However, AV-first cases, different from the other two types, don't have a pre-conflict duration shorter than \SI{2}{\second}. In terms of MRCT, AV-second cases show a significantly larger mean value. Recall that pre-conflict duration depends on the decision-making of the first-passing vehicle and the PET is mainly decided by the second-passing vehicle. Thus, both shreds of evidence suggest that in this dataset, AVs drive more conservatively and their capability of conflict resolution is less efficient than human drivers. Another observation is that when HV passes first, the pre-conflict duration is the same whether it interacts with an AV or another HV. Combined with the similar distributions of PET when HV passes the second (interacts with an HV or an AV), we argue that, on average, HV does not react differently to the presence of AV throughout the conflict resolution process.

MRCT is the time interval between passing two succeeding vehicles on the same stream. Therefore, 1/MRCT can be regarded as an indication of traffic flow (number of vehicles passing the conflict point per lane in a unit of time) under the three assumptions. The bottom plot of Fig.\ref{fig: efficiency} shows the average flow and the standard variance of the 3 scenarios. Quantitatively, compared to HV-HV interactions, the average efficiency of AV-first cases and AV-second cases decreases by 4.3\% and 13.1\%, respectively. If AV-first and AV-second have the same probability of happening, the average loss of efficiency is around 8.6\%. Note that this is not equivalent to the real traffic flow or intersection capacity. In practice, the alternative passing assumption (A-1) does not hold, different serving times for any two conflicting streams must be considered, and the penetration rate of AV also plays a critical role. This ``flow'' only quantitatively measures the efficiency of microscopic interaction by such a macroscopic flow.

\subsection{Findings}

We summarize the main findings of this section as follows:
\begin{itemize}
    \item By analyzing the surrogate safety measures and the interaction efficiency metric, subject to the Argoverse-2 dataset, \textit{HVs do not react differently to the presence of AVs in the conflict resolution} process at intersections.
    \item Regarding surrogate safety measures, a portion of \textit{pedestrians behave more conservatively when interacting with AVs} compared to interacting with HVs.
    \item On average, the AVs in Argoverse-2 drive more conservatively than HVs, which leads to potentially \textit{improved driving safety but unsurprisingly decreased traffic efficiency}. 
\end{itemize}

\section{Conclusion}
\label{sec: conclusion}

In this paper, we derive a high-quality conflict resolution dataset from the open Argoverse-2 data. This well-labelled dataset comprises diverse and balanced AV-involved and AV-free conflict resolution behaviours at intersections. The safety and efficiency impacts of autonomous vehicles are assessed based on the derived dataset. The primary analysis suggests that HVs do not behave differently when interacting with an HV or an AV during conflict resolution. Nevertheless, some pedestrians demonstrate more conservative behaviours when facing an AV than an HV. For the AVs in the Argoverse-2 dataset, their conservative driving strategy potentially improves traffic safety, but at the cost of traffic efficiency. In this paper, we also introduce a novel measure for conflict resolution efficiency, the Minimum Recurrent Clearance Time (MRCT). 

Based on these conclusions, we further propose several research directions. First, at unsignalized intersections, vulnerable road users, such as pedestrians, are sensitive to the presence of AVs. Therefore, the AV-pedestrian interaction cannot be completely learned from HV-pedestrian interaction data. How to improve the safety of AV-pedestrian interaction needs more attention than AV-HV interactions. Second, the assessment of the efficiency impact of AVs has not drawn as much attention as the functional safety of AVs. developing novel methods and standards to evaluate the traffic efficiency of mixed traffic flow from field test data is critical for the smooth development of AV in mixed traffic flow. We believe that this presented dataset can help address these concerns.

\section*{Acknowledgements}
This research is sponsored by the NWO/TTW project MiRRORS with grant agreement number 16270 and the TU Delft AI Labs program. We thank them for supporting this study.

\bibliographystyle{unsrt}  
\bibliography{references.bib}  


\end{document}